\documentclass[letterpaper, 10 pt, conference]{ieeeconf}  

\IEEEoverridecommandlockouts                              

\usepackage{graphicx}
\usepackage{xprintlen}
\usepackage{times}
\usepackage[utf8]{inputenc}
\usepackage{epsfig}
\usepackage{graphicx}
\usepackage{xcolor,colortbl}
\usepackage[labelformat=simple]{subcaption}

\usepackage{amssymb}
\usepackage{amsmath}
\usepackage{pifont}
\usepackage{array}
\newcolumntype{P}[1]{>{\centering\arraybackslash}m{#1}}
\usepackage{bbold}
\usepackage{nicefrac}
\usepackage{tikz}
\usetikzlibrary{positioning}
\usepackage{tikzscale}
\usepackage{pgfplots}
\pgfplotsset{compat=newest}
\usepackage[mode=buildnew]{standalone}
\usepackage{dsfont}
\usepackage{csquotes}
\usepackage{booktabs}
\usepackage{multirow}
\usepackage{makecell}
\usepackage{siunitx}
\usepackage[font=small]{caption}
\newdimen\owntablesep
\newcommand{\cmark}{\ding{51}}%
\newcommand{\putindex}[3]{\vtop{\hbox{\hspace{#3} $#1$}
            \hbox{\raise 6mm \hbox{$\scriptscriptstyle #2$}}}}

\newcommand{\gradx}[0]{\vtop{\hbox{\rm grad}
            \hbox{\raise 2.5mm \hbox{\rm \hspace{2mm} \footnotesize x}}}}

\newcommand{\grady}[0]{\vtop{\hbox{\rm grad}
            \hbox{\raise 2.5mm \hbox{\rm \hspace{2mm} \footnotesize y}}}}

\newcommand{\grad}[1]{\vtop{\hbox{\rm grad}
            \hbox{\raise 2.5mm \hbox{#1}}}}

\newcommand{\stz}{\rule{0mm}{2.3ex}}

\newcommand{\btb}{     \begin{tabbing}             }
\newcommand{\bte}{     \end{tabbing}               }

\definecolor{tu0}{rgb}{0.7451, 0.1176, 0.2353}

\definecolor{tu1}{rgb}{1.0000, 0.8039, 0.0000}
\definecolor{tu11}{rgb}{1.0000, 0.8627, 0.3020}
\definecolor{tu12}{rgb}{1.0000, 0.9020, 0.4980}
\definecolor{tu13}{rgb}{1.0000, 0.9412, 0.6980}
\definecolor{tu14}{rgb}{1.0000, 0.9608, 0.8000}

\definecolor{tu2}{rgb}{0.9804, 0.4314, 0.0000}
\definecolor{tu21}{rgb}{0.9882, 0.6039, 0.3020}
\definecolor{tu22}{rgb}{0.9882, 0.7137, 0.4980}
\definecolor{tu23}{rgb}{0.9922, 0.8275, 0.6980}
\definecolor{tu24}{rgb}{0.9961, 0.8863, 0.8000}

\definecolor{tu3}{rgb}{0.6902, 0.0000, 0.2745}
\definecolor{tu31}{rgb}{0.7529, 0.2000, 0.4196}
\definecolor{tu32}{rgb}{0.8431, 0.4980, 0.6353}
\definecolor{tu33}{rgb}{0.9216, 0.7490, 0.8196}
\definecolor{tu34}{rgb}{0.9529, 0.8510, 0.8902}

\definecolor{tu4}{rgb}{0.4863, 0.8039, 0.9020}
\definecolor{tu41}{rgb}{0.6431, 0.8627, 0.9333}
\definecolor{tu42}{rgb}{0.7412, 0.9020, 0.9490}
\definecolor{tu43}{rgb}{0.8431, 0.9412, 0.9686}
\definecolor{tu44}{rgb}{0.8980, 0.9608, 0.9804}

\definecolor{tu5}{rgb}{0.0000, 0.5020, 0.7059}
\definecolor{tu51}{rgb}{0.3020, 0.6510, 0.7961}
\definecolor{tu52}{rgb}{0.5490, 0.7765, 0.8667}
\definecolor{tu53}{rgb}{0.7490, 0.8745, 0.9255}
\definecolor{tu54}{rgb}{0.8510, 0.9255, 0.9569}

\definecolor{tu6}{rgb}{0.0000, 0.3255, 0.4549}
\definecolor{tu61}{rgb}{0.2510, 0.4941, 0.5922}
\definecolor{tu62}{rgb}{0.5490, 0.6941, 0.7529}
\definecolor{tu63}{rgb}{0.7490, 0.8314, 0.8627}
\definecolor{tu64}{rgb}{0.8510, 0.8980, 0.9176}

\definecolor{tu7}{rgb}{0.7765, 0.9333, 0.0000}
\definecolor{tu71}{rgb}{0.8431, 0.9529, 0.3020}
\definecolor{tu72}{rgb}{0.8863, 0.9647, 0.4980}
\definecolor{tu73}{rgb}{0.9333, 0.9804, 0.6980}
\definecolor{tu74}{rgb}{0.9569, 0.9882, 0.8000}

\definecolor{tu8}{rgb}{0.5373, 0.6431, 0.0000}
\definecolor{tu81}{rgb}{0.6784, 0.7490, 0.3020}
\definecolor{tu82}{rgb}{0.7686, 0.8196, 0.4980}
\definecolor{tu83}{rgb}{0.8588, 0.8941, 0.6980}
\definecolor{tu84}{rgb}{0.9059, 0.9294, 0.8000}

\definecolor{tu9}{rgb}{0.0000, 0.4431, 0.3373}
\definecolor{tu91}{rgb}{0.3020, 0.6118, 0.5373}
\definecolor{tu92}{rgb}{0.5490, 0.7490, 0.7020}
\definecolor{tu93}{rgb}{0.7490, 0.8588, 0.8353}
\definecolor{tu94}{rgb}{0.8549, 0.9176, 0.9059}

\definecolor{tu10}{rgb}{0.8000, 0.0000, 0.6000}
\definecolor{tu101}{rgb}{0.8706, 0.3490, 0.7412}
\definecolor{tu102}{rgb}{0.9216, 0.6000, 0.8392}
\definecolor{tu103}{rgb}{0.9608, 0.8000, 0.9216}
\definecolor{tu104}{rgb}{0.9804, 0.8980, 0.9608}

\definecolor{tu110}{rgb}{0.4627, 0.0000, 0.4627}
\definecolor{tu111}{rgb}{0.5961, 0.2510, 0.5961}
\definecolor{tu112}{rgb}{0.7294, 0.4980, 0.7294}
\definecolor{tu113}{rgb}{0.8392, 0.6980, 0.8392}
\definecolor{tu114}{rgb}{0.9216, 0.8510, 0.9216}

\definecolor{tu120}{rgb}{0.4627, 0.0000, 0.3294}
\definecolor{tu121}{rgb}{0.6118, 0.3020, 0.5333}
\definecolor{tu122}{rgb}{0.7569, 0.5490, 0.6980}
\definecolor{tu123}{rgb}{0.8667, 0.7490, 0.8314}
\definecolor{tu124}{rgb}{0.9216, 0.8510, 0.9020}

\definecolor{tu130}{rgb}{0.0314, 0.0314, 0.0314}
\definecolor{tu131}{rgb}{0.3725, 0.3725, 0.3725}
\definecolor{tu132}{rgb}{0.5882, 0.5882, 0.5882}
\definecolor{tu133}{rgb}{0.7529, 0.7529, 0.7529}
\definecolor{tu134}{rgb}{0.8667, 0.8667, 0.8667}

\title{\LARGE \bf
Corner Cases for Visual Perception in Automated Driving: \\Some Guidance on Detection Approaches
}

\author{Jasmin Breitenstein$^{\ast}$, Jan-Aike Termöhlen$^{\ast}$, Daniel Lipinski$^{\circ}$ and Tim Fingscheidt$^{\ast}$
	\thanks{$^{\ast}$Jasmin Breitenstein, Jan-Aike Termöhlen and Tim Fingscheidt are with the Institute for Communications Technology, Technische Universit{\"a}t Braunschweig, Schleinitzstr. 22, 38106 Braunschweig, Germany. Email: {\tt\small \{j.breitenstein, j.termoehlen, t.fingscheidt\}@tu-bs.de}}%
	\thanks{$^{\circ}$Daniel Lipinski is with \mbox{Volkswagen AG}, Berliner Ring 2, 38440 Wolfsburg, Germany. Email:	
		{\tt\small daniel.lipinski@volkswagen.de}}
}

\begin{document}

\maketitle
\thispagestyle{empty}
\pagestyle{empty}

\begin{abstract}

Automated driving has become a major topic of interest not only in the active research community but also in mainstream media reports. Visual perception of such intelligent vehicles has experienced large progress in the last decade thanks to advances in deep learning techniques but some challenges still remain. One such challenge is the detection of corner cases. They are unexpected and unknown situations that occur while driving. Conventional visual perception methods are often not able to detect them because corner cases have not been witnessed during training. Hence, their detection is highly safety-critical, and detection methods can be applied to vast amounts of collected data to select suitable training data. A reliable detection of corner cases will not only further automate the data selection procedure and increase safety in autonomous driving but can thereby also affect the public acceptance of the new technology in a positive manner.
In this work, we continue a previous systematization of corner cases on different levels by an extended set of examples for each level. Moreover, we group detection approaches into different categories and link them with the corner case levels. Hence, we give directions to showcase specific corner cases and basic guidelines on how to technically detect them.
\end{abstract}

\section{Introduction}

Automated driving and its technologies have experienced significant progress in the last years. While this progress has been made and advances in automated driving have received a lot of attention, it still faces some challenges for a safe and reliable application in daily life.  Visual perception methods form an important part of the intelligent vehicles. They are expected to detect, and understand their environment. 
Therefore, there already exists a vast amount of algorithms for visual perception tasks associated with the vehicle's environment including object detection (e.g., \cite{Ren2015}), semantic segmentation (e.g., \cite{Romera2018,Ronneberger2015}), instance segmentation (e.g., \cite{He2017}), and many more. However, one crucial factor is the behavior of visual perception methods in unexpected situations that deviate from  normal traffic situations. 
Those situations, so-called \emph{corner cases}, exist in an infinite number of examples. Their dominant and connecting feature is their deviation from what is generally considered normal traffic behavior. 
Possible corner cases are for example the classical situations everyone fears of encountering while driving, such as a person running onto the street from behind an occlusion, a ghost driver or simply lost cargo on the street.

\begin{figure}[htb!]\vspace*{0.15cm}
\centering
\hspace*{-0.2cm}
\resizebox{0.45\textwidth}{12.5cm}{%
\includegraphics[width=\textwidth]{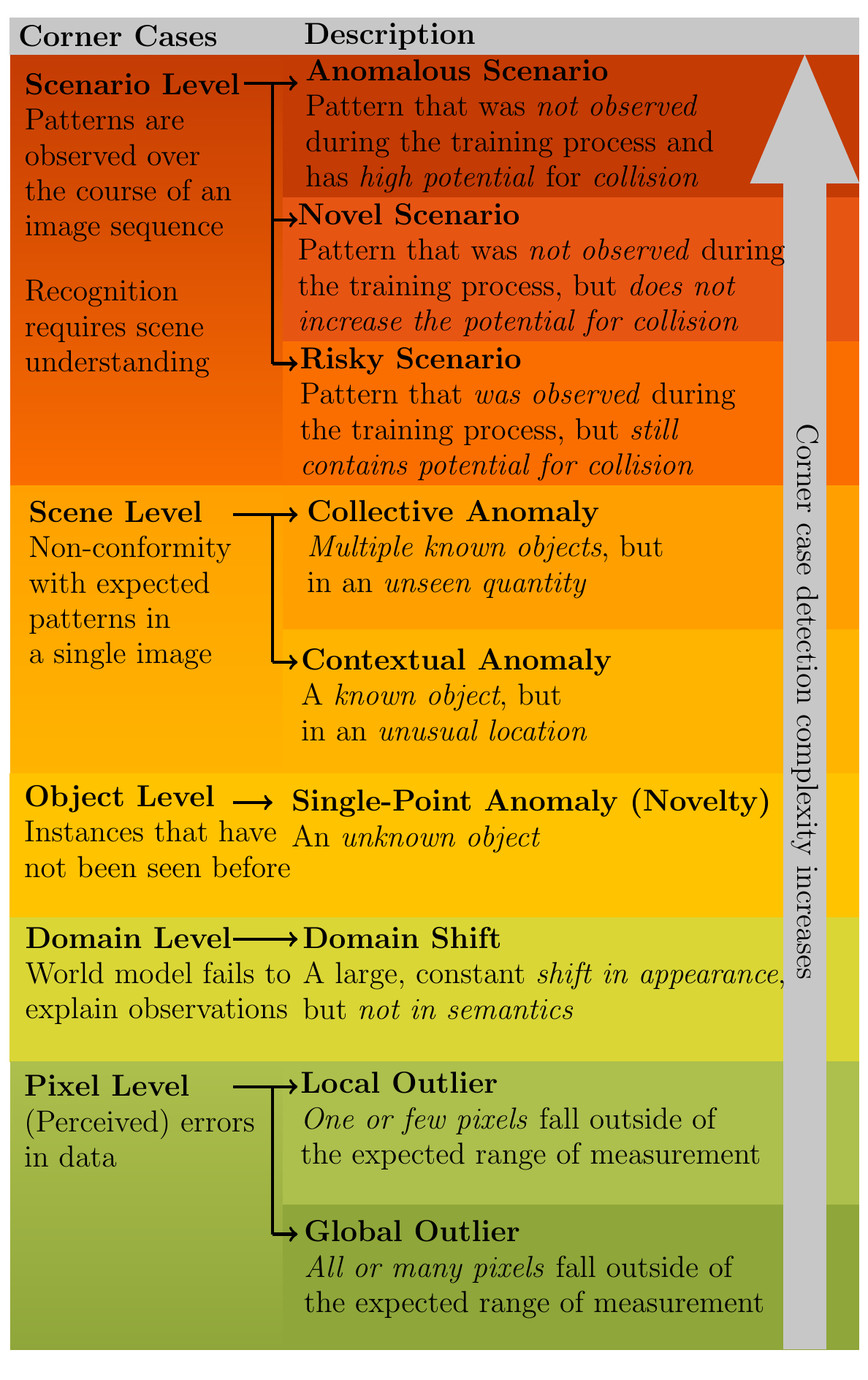}
     }
	\caption{Systematization of corner cases on different levels as given in \cite{Breitenstein2020}. The theoretical complexity of the detection typically increases from the bottom to the top.}
	\label{fig:short}
\end{figure}

A reliable detection of such corner cases is crucial for safety in automated driving as it can reduce the number of accidents of autonomous cars, and hence helping widespread acceptance and application of this technology. 
It is necessary both online, in the vehicle, and offline applications during development. 
A stable and confident corner case detection method recognizes critical situations.
In the online application, it can be used as a safety monitoring and warning system which identifies situations while they occur. 
In the offline application, the corner case detector is applied to large amounts of collected data to select suitable training and relevant test data in the development of new visual perception algorithms in the laboratory. 
While the detection both online and in the laboratory are related to safety, the offline applications also lead to both monetary and time savings by automatizing the selection process of training data. Hence, there already exists a variety of works dealing with the detection of corner cases in the automotive context such as the detection of obstacles \cite{Pinggera2016, Ramos2017}, or newly appearing objects \cite{Blum2019}.

Even though faithful and efficient corner case detection will have a large impact on automated driving, consistent generally accepted definitions and classifications are still missing to describe them. We follow the definition, that corner cases are present when ``there is a non-predictable relevant object/class in a relevant location'', of Bolte et al. \cite{Bolte2019b}. To facilitate the systematical development of detectors, a categorization was introduced in \cite{Breitenstein2020}. A trimmed version of the corner case systematization can be found in Figure \ref{fig:short}.
It depicts a hierarchy ordered by the theoretical complexity of detection.
We consider corner cases on pixel, domain, object, scene, and scenario level, which are described in more detail in Section \ref{sec:systematization}. 

While this systematization  already paves the way for a more methodical development of detection methods, it also poses the question on how to actually detect the specific corner cases of each level.
In the context of smart manufacturing, a systematization of possible faults of the system has been established by Lopez et al. \cite{Lopez2017}. Next to this systematization, the authors propose a categorization of detection methods into feature extraction, regression, knowledge-based, signal model, state estimation, clustering, and classification methods, connecting each method category with specific anomaly categories in smart manufacturing. We follow their example in order to extend the previous systematization of corner cases for visual perception in automated driving by a categorization of detection approaches, and associating them with the previously defined corner case levels. Additionally, we give both specific examples of corner case for each level and provide first guidelines towards basic detection methods.

Due to the omnipresence and success of deep learning methods in visual perception algorithms, we limit this categorization to deep learning methods. 
They are highly functioning methods with promising results in many visual perception applications and have also shown success in the detection of unusual occurrences. Moreover, we limit this work to purely visual approaches excluding other sensor data such as RADAR and LiDAR, but consider corner cases that can be detected either from single image frames or entire image sequences.  

The paper is structured as follows. We briefly review the systematization of corner cases in Figure \ref{fig:short}. Then we provide more thorough examples for each corner case level with the purpose of enabling both a more comprehensive understanding of what kind of situation the corner case levels can encompass, and the recording of corner cases by almost mimicking stage directions.
Moreover, we extend the previous systematization by categories for detection approaches with their respective related work. Finally, we map detection approaches to corner case levels by providing hints and intuition for the development of new methods.


\section{Systematization of Corner Cases}
\label{sec:systematization}

Previously, a systematization of corner cases for visual perception in automated driving has been introduced \cite{Breitenstein2020}, which we briefly summarize in the following. This systematization can be found in Figure  \ref{fig:short} in a reduced version. The corner case levels are ordered by detection complexity. Going from lower to higher complexity of detection, we have corner cases on pixel level which can be divided into \textbf{global} and \textbf{local outliers}. Examples for those are overexposure and dead pixels, respectively. Then there are domain-level corner cases which are caused by \textbf{domain shifts} such as for example a change in location, weather, or the time of the day. On object level, corner cases are \textbf{single-point anomalies} or novelties. This can for example be a wild animal, e.g., a lion, appearing on the street, or walking aids such as a rollator, or crutches. For scene-level corner cases, we again distinguish between two types: collective and contextual anomalies. \textbf{Contextual anomalies} denote known objects in unusual locations as a tree in the middle of the street. \textbf{Collective anomalies} are known objects in unusual quantities such as a demonstration. 

The highest complexity of detection have scenario-level corner cases, which are observed over the course of an image sequence. \textbf{Risky scenarios} have been observed in a similar fashion but still pose a potential for collision such as overtaking a cyclist.
\textbf{Novel scenarios} have not been observed but do not increase the potential for collision such as accessing the freeway. \textbf{Anomalous scenarios} also have not been observed, but pose a very high potential for collision such as a person suddenly walking onto the street in front of the ego vehicle.

While in the introduction of the systematization \cite{Breitenstein2020}, the different corner case levels were discussed in detail and suitable datasets and metrics were pointed out, in Section \ref{sec:detection} we extend this systematization by another dimension, following the approach of Lopez et al.\ \cite{Lopez2017}. In this dimension, different detection methods are grouped into broad categories and linked to the respective corner case levels. Moreover, we extend the columns in Figure \ref{fig:short} by a comprehensive list of examples which basically provides a playbook of corner cases.
%


\section{Showcasing Corner Cases}
\label{sec:playbook}

In Table \ref{tab:playbook-coca}, we provide examples for the corner case levels in Figure \ref{fig:short}. This is supposed to clarify what corner cases can be found on each level of the systematization and serve almost as stage directions for possible corner case recordings.
Moreover, it also gives an indication as to what content of datasets is needed to develop and test reliable corner case detectors. Again, Table \ref{tab:playbook-coca} sorts the example situations by their respective corner case levels. The situations are described in some detail, so they can be directly translated into directions for data acquisition. Furthermore, the following sections will introduce categories of detection methods to later associate with respective corner case levels and thus, give some guidelines to detect the example corner cases depicted in Table \ref{tab:playbook-coca}.

\begin{table*}[htp!]
\begin{center}
\begin{tabular}{|P{1cm}|P{1.5cm}|P{12.5cm}|}
\hline
\multicolumn{2}{|c|}{\stz Corner Case Level} & Example  \\
\hline
\parbox[t]{2mm}{\multirow[c]{3}[6]{1cm}{\rotatebox[origin=c]{90}{\hspace*{-4.0cm}Scenario Level}}}
&\stz Anomalous Scenario \; (potentially dangerous, unknown) & 
\begin{itemize}
\item We are driving through a street. On our right side, a large construction trailer is parked at the roadside. While we are driving next to it, suddenly a person steps onto the road. Before appearing on the road in front of the trailer, the person was fully occluded by it.
\item We are driving through a busy road with multiple lanes. Everyone is driving relatively fast. On the lane to our left, there is another car driving. It is slightly in front of us. Suddenly, the car changes into our lane without giving any indication first and it is causing us to brake heavily in order to avoid a collision.
\item We are driving in a street and are approaching a crossroad. The traffic signs indicate that we have the right of way, also indicated by stop signs on the other street. Due to the surrounding houses, we cannot see into the other street approaching the crossroad. While we drive onto the crossroad, another traffic participant does not abide to the traffic rules and is not stopping at the stop sign. They drive directly onto the crossroad at the same time as we do, not yielding our right of way.
\end{itemize} \\
\cline{2-3}
 &\stz Novel Scenario \;   (no potential danger, unknown) & 
\begin{itemize}
\item We drive towards a railroad crossing. The visual perception algorithm has not seen this before. The gates at this crossing are open. Thus, it is safe to cross.
\item We are driving to another city and want to use the freeway to get there. While the training data contained changing lanes and turnings, it did not contain the specific situation of accessing the freeway. 
\item We are driving through a residential area in an urban environment. It is difficult to find parking spots in this area. We see a small car at the side of the road. Driving towards it, we first believe that it is driving onto the street, but closing in, we realize that it is actually parking orthogonally to the other cars to fit into an even smaller parking spot. This way of parking, we first interpreted as a driving car because of its positioning orthogonal to the street.

\end{itemize} 
  \\
 \cline{2-3}
 &\stz Risky Scenario \; (potentially dangerous, known) &  
 \begin{itemize}
 \item We drive through a narrow street. Another car is driving towards us. As the cars pass each other, there is not much space left between them.
 \item We drive through a street. While there is a sidewalk, there is no separate cycling lane. Driving further, there is a cyclist driving on the street. As the car is faster than the cyclist, we overtake them.
 \item We drive through a street with only one lane for each direction. The motorized vehicle in front of us is going considerably slower than the allowed speed limit. When there is no oncoming traffic, we overtake the other vehicle by driving on the lane for the other direction.
 \end{itemize} \\
 \hline \hline
 \parbox[t]{2mm}{\multirow[c]{2}[4]{1cm}{\rotatebox[origin=c]{90}{\hspace*{-1.3cm}Scene Level}}}
&\stz Collective Anomaly &  
\begin{itemize}
\item In the city center, there is a demonstration and many people are walking and standing on the street, holding banners and shouting paroles.
\item It is in the evening and most people have left work and are on their way home. It is rush hour and we incur a major traffic jam.
\item We are at a crossroad. While it is usually governed by traffic signs, at the moment there is a huge construction site. Next to the usual signs, there are now many more signs to govern the behavior around the construction site.
\end{itemize}
\\
\cline{2-3}
 &\stz Contextual Anomaly & 
\begin{itemize}
\item We are in the city. The night before there was a huge storm and a tree fell on the street.
\item On a street, oil was spilled. While this is no longer visible, there are still traffic cones on the street, indicating to drive around the spot where it happened.
\item A car is parking on the sidewalk in a street with only few available actual parking spots.
\end{itemize} 
  \\
 \hline \hline
 \rotatebox[origin=c]{90}{\hspace*{0.35cm}Object Level} &\stz Single-Point Anomaly & 
\begin{itemize}
\item Unexpectedly, in a residential area, there is a bear in the middle of the street.
\item At a traffic light, a person with a rollator and a person on crutches cross the street.
\item It is a sunny day and we can see a shadow of a pedestrian approaching our street to cross, but the actual pedestrian is still occluded by a wall or the roadside.
\end{itemize} 
  \\
 \hline \hline
 \rotatebox[origin=c]{90}{\hspace*{0.7cm}Domain Level} &\stz Domain Shift &  
\begin{itemize}
\item It is February and it has snowed heavily the past couple of days. We are driving through the city and at the side of the road there are piles of snow everywhere.
\item We are going on vacation and drive to Great Britain by car all the way from continental Europe. After passing the Channel Tunnel, everyone is driving on the left.
\item We usually live in a city and the visual perception system has only been trained on urban data. On a sunny spring day, we decide to take a trip to the rural surrounding areas.
\item We are driving through a busy street in the city. It is rainy and dark. Moreover, there is a lot of oncoming traffic.
\end{itemize}  
    \\
 \hline \hline
 \parbox[t]{2mm}{\multirow[c]{2}[3]{1cm}{\rotatebox[origin=c]{90}{\hspace*{-1.3cm}Pixel Level}}}
 &\stz Local Outlier & 
\begin{itemize}
\item Our camera fell down and now there are dead or broken pixels.
\item It is a windy day and there appears dirt on the windshield.
\item It is fall and a leaf has fallen onto our windshield.
\end{itemize} 
 \\
\cline{2-3}
 &\stz Global Outlier & 
\begin{itemize}
\item We are driving in a tunnel. It is a sunny day and at the end of the tunnel there is overexposure on our camera images because of the sun.
\item We drive at night and another car is coming towards us. The lights of this car have not been adjusted properly and we are blinded by them.
\item We are driving through a street during sunset. When we turn into another street leading west, we are blinded by the setting sun.
\end{itemize} 
 \\
 \hline 
\end{tabular}
\caption{Example situations on each level of the corner case systematization as shown in Figure \ref{fig:short}.}
\label{tab:playbook-coca}
\end{center}
\end{table*}

\section{Concepts of Detection Approaches}
\label{sec:detection}

We distinguish between five broad concepts of detection approaches: reconstruction, prediction, generative, confidence scores, and feature extraction. We subdivide the confidence score category into learned scores, Bayesian approaches and scores obtained by post-processing.  

\textbf{Reconstruction} approaches are typically based on autoencoder-type networks. Most of those methods follow the paradigm that normality can be reconstructed more faithfully than anomalies. This causes reconstruction-based approaches to appear on every level of the corner case hierarchy. Especially, they can be applied to both the corner cases concerning single images and those comprising an entire image sequence.
Hasan et al.\ \cite{Hasan2016} train an autoencoder both end-to-end and on handcrafted features where the reconstruction error serves as anomaly score. While they consider video sequences, there also exist similar approaches with single images as network input \cite{Xia2015}. 
Some reconstruction approaches rely on prototypical learning. During training, prototypes of the normal samples are learned in the latent space, which during inference lead to a more faithful reconstruction of normal samples compared to anomalous ones \cite{Gong2019a}. Oza et al.\ \cite{Oza2019} also utilize reconstruction in a class-conditioned autoencoder for open-set recognition, where next to the unsupervised open-set training they perform supervised closed-set training.

\textbf{Prediction-based} approaches can be mostly found on scenario level. Typically, they predict a future frame and later compare it with the true frame to detect any anomalies. Thus, they can be trained in a supervised manner where we assume that all training samples are normal. Such a method has been applied by Bolte et al. \cite{Bolte2019b} specifically for corner case detection in automated driving. Another approach predicts future frames in videos using a generative adversarial network structure while ensuring appearance and motion constraints \cite{Liu2018}.

\textbf{Generative} and reconstruction-based approaches are closely related, since also this type of method can base its decision on the reconstruction error. Generative approaches, however, also regard the discriminator's decision or the distance between the generated and the training distribution. Moreover, some also borrow simply from related techniques such as adversarial training or perturbations. 
Lee et al.\ \cite{Lee2018} introduce a confidence loss to enforce low confidence on out-of-distribution samples while also generating boundary out-of-distribution training samples for this task and jointly train the generative objective with a classification one. 
Adversarial-based training is also used to enforce uniform confidence predictions on noise images, leading simultaneously to decreased confidence for anomalous samples \cite{Hein2019}.
Based on generating images from masks, Lis et al.\ \cite{Lis2019} identify unknown objects in the data by considering the error between the generated image and the original one.
Generative methods, namely variational and adversarial autoencoders, have been applied to collective anomaly detection considering the error between the original and the generated image as anomaly score \cite{Chalapathy2019}. 
Also, domain shifts can be measured by generative methods. For example, representation learning guided by the Wasserstein distance \cite{Shen2018} uses a general adversarial network inspired architecture, where a domain-critic network estimates the Wasserstein distance between source and target domain features. L\"ohdefink et al.\ \cite{Loehdefink2020} apply a generative adversarial network based autoencoder to detect domain shifts based on the earth mover's distance \cite{Rubner2000}.

Methods relying on \textbf{confidence scores} are grouped into three categories to provide more clarity for this approach. Those that obtain their scores by post-processing, those that learn their scores, and those that rely on Bayesian approaches.

Confidence scores can be based on applying \textit{post-processing} techniques to the neural networks without interfering with the training process.
A baseline approach exists obtaining confidence scores by comparing softmax values against fixed thresholds \cite{Hendrycks2017}.
Moreover, the scores are, for example, obtained by applying a Kullback-Leibler divergence matching to the softmax outputs during inference to compare with the class-specific templates obtained from a validation set in a multi-class prediction setting \cite{Hendrycks2019}. The method is trained in a supervised manner without requiring anomalous examples and gives segmentation maps as output.  
Another post-processing approach employs temperature scaling to the inputs \cite{Liang2018}. It is based on the paradigm that for such modified normal inputs, the network is still able to infer the correct class but not for unknowns. It is also only supervised via normal training samples.

In contrast to obtaining confidence scores by post-processing, they can also be learned during training. In this category of \textit{learned confidence scores}, we also include any method that relies on the training set in general to, e.g., provide a threshold.
As an example for thresholds based on the training set, Shu et al. \cite{Shu2018b} compute three thresholds: one for acceptance of a sample, one for rejection and a distance-based one for when a sample falls in between the two other thresholds. As the threshold-based values are based on the training set, we consider the resulting confidence scores as learned. While the method is trained in a supervised manner, no corner case examples are required during training. The method then outputs a label as either one of the known classes or as unknown. 
Another way to obtain learned confidence scores is to borrow techniques from multi-task learning and include a second branch into the network to learn confidence scores \cite{Devries2018}. Also in this case, training is done in a supervised manner, but only requiring normal samples. While originally used for classification, this method has been adapted to segmentation as well \cite{Hendrycks2019}. Learned confidence scores can be obtained by prototypical learning, where scores are based on the distance to normal training samples \cite{Xing2020}. The open-max activation also provides learned confidence scores after a supervised training with normal samples with the aim to detect unknowns \cite{Bendale2016}.
Learning to detect geometric transforms applied to the data from normal training data, also leads to learned confidence scores by assuming that these transforms can be more accurately detected on normal samples \cite{Golan2018}.

\textit{Bayesian confidence scores} are usually obtained by an estimation of the model uncertainty, the epistemic uncertainty \cite{Kendall2017}. A network is trained to output a posterior distribution over its weights. Typical examples of this type of methods include the Monte Carlo dropout technique \cite{Kendall2017,Gal2016} or deep ensembles \cite{Lakshminarayanan2017}. The supervised training relies on normal training data.
A current method to obtain model uncertainty scores is for example deterministic uncertainty quantification, which is based on ideas from radial basis networks \cite{vanAmersfoort2020}.
In the semantic segmentation setting, Bayesian neural networks provide both class labels and an estimation of the model's uncertainty about it for each pixel. Common measures for this uncertainty include entropy and variance \cite{Gal2016}. 
An example for the application of Monte Carlo dropout for uncertainty estimation in semantic segmentation is introduced by the Bayesian SegNet, which includes dropout units into the network architecture to obtain confidence maps for the model \cite{Kendall2015}. 
Pham et al. \cite{Pham2018} use a Bayesian framework for instance segmentation in an open set recognition setting.
An extension to include an entire time span has been achieved by considering the moving average over multiple frames \cite{Huang2018c}.

\textbf{Feature extraction} approaches employ deep neural networks to extract features from the input data. These features are then either further processed using another technique or they are directly used to provide a classification label. 
In contrast to confidence scores, feature extraction methods either directly classify the sample as corner case or they use the extracted features in another way to obtain their decision. Confidence scores typically provide the scores next to their decided label.
One such approach extracts features which are then fit on a hypersphere during training \cite{Ruff2018}. Thus, while the approach is unsupervised, it requires the training data to be considered normal because in inference, data is decided as anomalous if the distance on the hypersphere exceeds a threshold. When regarding video sequences, features can also be extracted from single frames and then be considered over a particular time interval to compare the probability distribution in the interval with the one outside \cite{Barz2019}. 
Classification-reconstruction learning for open-set recognition (CROSR) also learns a feature representation for an unknown class detector \cite{Yoshihashi2019}. The feature representation is composed of latent representations learned from reconstructing each intermediate layer of the network. The class membership is modeled via an extreme-value-theory-based distribution of the distances between extracted features of normal training data and the respective class mean \cite{Yoshihashi2019}.
Standard classification approaches train a network in a supervised manner, using the softmax function as activation on the last layer to obtain a corresponding class for the input sample. Jatzkowski et al.\ \cite{Jatzkowski2018} utilize such an approach for overexposure detection.
Feature extraction approaches are also found in domain adaptation methods. There, for example, cross-entropy based metrics \cite{Zou2018a,Chen2018a} are minimized in the adaptation process, indicating that they are valid measures of domain mismatch. Bolte et al.\ \cite{Bolte2019a} consider the mean-squared error of extracted features as a measure of domain shift.

\begin{table*}[htb!]
\begin{center}
\begin{tabular}{|P{1cm}|P{1.5cm}|P{1.5cm}|P{1.5cm}|P{1.5cm}|P{1.5cm}|P{1.5cm}|P{1.5cm}|P{1.5cm}|}
\hline
\multicolumn{2}{|c|}{ \multirow[c]{2}[4]{2.3cm}{\stz Corner Case Level}}& \multirow[c]{2}[4]{1.5cm}{\stz Prediction}& \multirow[c]{2}[4]{1.5cm}{\stz Reconstruction}&\multirow[c]{2}[4]{1.5cm}{\stz Generative}&\multirow[c]{2}[4]{1.5cm}{\stz Feature Extraction}&\multicolumn{3}{|c|}{\stz Confidence Score}\\
\cline{7-9}
\multicolumn{2}{|c|}{} &  &  &  &   & Post-processing & Bayesian & Learned  \\
\hline
\multirow[c]{3}[6]{1cm}{\centering Scenario Level} &\stz Anomalous Scenario & \cmark$^\star$&\cmark&&\cmark&\cmark$^\star$&\cmark& \\
\cline{2-9}
 &\stz Novel Scenario &  \cmark$^\star$
&\cmark&&\cmark&\cmark$^\star$&&
  \\
 \cline{2-9}
 &\stz Risky Scenario &  \cmark$^\star$
&\cmark&&\cmark &\cmark$^\star$&\cmark & \\
 \hline \hline
\multirow[c]{2}[4]{1cm}{\centering Scene Level} &\stz Collective Anomaly &  
&&\cmark&&\cmark$^\star$&&
\\
\cline{2-9}
 &\stz Contextual Anomaly & 
&&&\cmark &&\cmark$^\star$&
  \\
 \hline \hline
 Object Level &\stz Single-Point Anomaly & 
&&\cmark$^\star$& \cmark &\cmark$^\star$&\cmark$^\star$&\cmark$^\star$
  \\
 \hline \hline
  Domain Level &\stz Domain Shift &  
&& \cmark$^\star$ & \cmark &&& 
    \\
 \hline \hline
 \multirow[c]{2}[3]{1cm}{\centering Pixel Level} &\stz Local Outlier & 
\cmark$^\star$ &&&&&&
 \\
\cline{2-9}
 &\stz Global Outlier & 
&&&\cmark &&&
 \\
 \hline 
\end{tabular}
\caption{Detection approaches are ascribed to corner case levels. A $\star$ denotes the suggested approaches to detect corner cases on that level in Section \ref{sec:mapping}.}
\label{tab:mapping}
\end{center}
\end{table*}

\section{Associating Detection Approaches and Corner Case Levels}
\label{sec:mapping}

In this section, we associate the detection approaches of Section \ref{sec:detection} with the corner case levels in Figure \ref{fig:short}, as was similarly done for smart manufacturing \cite{Lopez2017}. We discussed some examples for each detection concept in Section \ref{sec:detection}, which already hint at which type of corner case they can be suitably applied to. Moreover, we wish to prompt an idea on how to detect certain corner cases, as for example the ones listed in Table \ref{tab:playbook-coca}. A summary of this section can be found in Table \ref{tab:mapping}, where we denote, which type of method has been applied to detect which corner case level. Additionally, we point out which approaches we believe to be leading to efficient and reliable future detection methods.

Overall, it can be said that due to the lack of large scale datasets containing all types of corner cases, and the associated open-world problem of corner case detection, unsupervised methods or ones trained only on normal samples currently seem to be the most effective way to obtain corner case detectors. Approaches dependent on anomalous training data need a more complex and specialized training set and run the risk of focusing on the specific corner cases related to its samples, thus turning a blind eye on the possibility of unknown corner cases appearing in inference.

On \textit{pixel level}, to our knowledge only few deep learning approaches exist. But to detect such corner cases, for global outliers, feature extraction approaches provide promising results \cite{Jatzkowski2018}, since we aim to detect a corner case which influences large parts or even the entire image. In this case, the detection can be considered as a binary classification problem and a network is able to extract sufficient features for that task. Supervised training is possible because there is no unexpected amount of diversity for this type of corner case. Due to the lack of datasets for automated driving with labeled global outliers such as overexposure, however, an investigation of methods utilizing few-shot learning or similar techniques could be beneficial. Moreover, we are interested in the detection of multiple global outliers such as, for example, detecting overexposure and underexposure in images jointly. In the case of exiting a tunnel, they can even appear in the same image. This can be investigated in future work by considering joint or multi-task-learning.

Local outliers influence only a small portion of the image, as in the case of dead pixels. Detection of those corner cases can be learned supervisedly, as it can be simulated in the training data. Due to the possibility of simulation, detection could be treated in a semantic segmentation method by including another class. This will also lead to pixel-wise labels which will inform about the location of the dead pixels.
We believe detection of local outliers will profit from taking a predictive approach, and thus including a time span. A predicted location of, for example, dead pixels can be compared to the actual location. Ideally, the actual location is in contrast to the predicted one based on the learned optical flow.


To detect \textit{domain-level} corner cases, we do not need to use domain adaptation methods but find suitable measures of domain mismatch. These measures, however, often stem from domain adaptation methods where they are utilized as loss functions.
Typically, such measures are considered feature extraction approaches. While the training can require supervision by normal samples from a source domain, data from another domain for training should be explicitly excluded. Methods that employ specific examples of a second domain in training are in danger not to reach the same performance for a third domain. Bolte et al.\ \cite{Bolte2019a} use the mean-squared error distance to measure the difference between features in the source and target domain in an unsupervised domain adaptation setting. 
It could also prove advantageous to consider out-of-distribution detection methods that are typically evaluated by considering one dataset as in- and another one as out-of-distribution. Out-of-distribution detectors are often evaluated by distinguishing between a dataset they have been trained on and another one \cite{Hendrycks2017,Lee2018}. Those methods could be extended from a classification setting to the automotive visual perception setting, as they only require training supervised via normal samples.  
For reliable detection of domain-level corner cases, we require trustworthy measures of domain mismatch. To that end, we depend upon an evaluation using more than just one target domain. One such measure applying a generative adversarial network based autoencoder has been introduced previously providing a new domain mismatch metric based on the earth mover's distance \cite{Loehdefink2020}.

On \textit{object level}, the main goal is to detect \textit{unknown unknowns} \cite{Scheirer2014}. These are instances belonging to a new class not seen before in training. Providing examples of such corner cases during training would lead the network at inference to detect only corner cases similar to those examples, which is self-defeating to our aim. Detection of object-level corner cases falls into the broad area of open-set recognition and the related approaches usually provide some type of confidence score. Ideally, for detection and localization we ask for pixel-wise scores. There also exist reconstruction and generative methods which comply with this idea. However, reconstruction-based approaches tend to provide less meaningful results \cite{Lis2019}. We would like to obtain a semantic segmentation mask for the input images where the pixels belonging to unknown objects are associated with an unknown class label or with a high amount of prediction uncertainty. With this goal in mind, pursuing confidence score and generative detection methods seems the most fruitful, and many recent methods comply with this indication \cite{Hendrycks2019,Lis2019,Xing2020}. Using Bayesian confidence scores, we ask for a model with high uncertainty associated to those unknown objects. Here, scalable methods for Bayesian deep learning applying Monte-Carlo dropout \cite{Gal2016} or deep ensembles \cite{Lakshminarayanan2017} provide a first step towards detection. In terms of the definition of those single-point anomalies as unseen instances during training, we conjecture that efficient and reliable detection approaches cannot rely on training samples including corner cases. Here, one has to resort to unsupervised approaches which can merely be trained using samples of normality. 

On \textit{scene level}, we aim to detect known classes in either unseen quantities or locations. Chalapathy et al. \cite{Chalapathy2019} employ generative methods to detect collective anomalies, which achieves promising results. Moreover, we believe future work should leverage instance segmentation to obtain a group size by counting the number of instances of each class. In this case, a threshold is required for defining a collection as anomalous.
Detecting contextual anomalies can be approached by employing feature extraction approaches \cite{Ruff2018}. However, in the case of automotive visual perception, feature extraction might not be able to capture the complexity of the entire scene.
Hence, many existing methods give confidence scores \cite{Kendall2015,Lakshminarayanan2017} or reconstruction errors \cite{Gong2019a} and distinguish between normal and anomalous samples. 
We propose to investigate how an incorporation of class priors influences the process as those priors might be able to facilitate the detection of misplaced class representatives. In the same light, confidence scores resulting from Bayesian deep learning indicate where the model is uncertain, and hence they can be useful to localize objects appearing in an unusual context. Both corner case types on scene level can be trained supervisedly with normal data since both detect instances of known classes, just either in an unusual location or quantity. However, in contrast to object-level corner cases, in this case while we might want pixel-wise semantic segmentation labels for our visual perception application, we additionally ask for instance-wise labels telling us an object appears in an unusual location, or image-wise labels if there appears an unseen quantity.

\textit{Scenario-level} corner cases are comprised of patterns that appear over a particular time span and might not seem anomalous in a single frame. Here, prediction-based methods whose decision depends on the comparison between a predicted and the actual frame provide rewarding results \cite{Bolte2019b}. Purely reconstructive methods obtain again less faithful corner case detection scores. 
Predictive approaches can be trained supervisedly, as they only require normal training samples in order to detect corner cases during inference. This is especially important for novel and anomalous scenarios, where we cannot capture every possibility due to the infinite number and the considerable danger of the corresponding circumstances. Also, including samples could actually prejudice the network to only detect such scenarios.  
For future work, we need to define adequate metrics for detection of this type of corner case. While we might still want to know the location of the corner case in the image, we also require the point of time when such a corner case happens. To achieve this, we could consider image-wise labels over a certain time span. 
Next to an investigation of metrics, we suggest the use of a cost function to give higher priority to the detection of vulnerable road users appearing at the verge of the visual field. This could improve, for example, the detection of a person running onto the street from behind an occlusion, as the person could already detected when just a few human pixels appear in the frame. Such an approach also requires a frame-wise mask identifying pixels which have not been included in the previous one because they have been occluded or outside the field of view.
  
While the detection approaches for all corner case categories have been treated separately, we also need to discuss the concept of a general corner case metric. Considering, there already is an ideal corner case detector available, that we want to apply to select training data for a visual perception module. It takes as input an entire video sequence with all types of corner cases contained in it, the question poses how to report the results. While we, e.g., suggest to report pixel-wise labels on object level but image-wise ones on pixel level, it needs to be clarified that in the end this needs to be combined to a general metric, which expresses if a video sequence contains corner cases, and thus qualifies as necessary training data. Here, one could consider a type of averaging metric similar to common velocity measures.


\section{Conclusions}

After reviewing the systematization of corner cases, we introduced a more detailed list of examples meant for deeper understanding of the previously proposed categories and enabling direct application for the acquisition of corner case data. Moreover, we extended the corner case systematization by covering detection approaches and their respective categories. Afterwards, we associated the detection approaches to corner case levels, and additionally provided some basic guidelines on how to detect certain types of corner cases. Hence, we are able to portray specific corner case examples and follow coarse guidelines for a baseline detection approach. 



\section*{Acknowledgment}
The authors gratefully acknowledge support of this work by Volkswagen AG, Wolfsburg, Germany.

{\small
	\bibliographystyle{IEEEbib}
	\bibliography{../bibliography/ifn_spaml_bibliography}
}

\end{document}